\documentclass[11pt]{article}

\usepackage{arxiv24-fine-granular-trigger-warnings}

\graphicspath{{../arxiv24-fine-granular-trigger-warnings-figures},{.}}

\begin{document}
\title{If there's a Trigger Warning, then where's the Trigger?\\Investigating Trigger Warnings at the Passage Level}

\newcommand{\buw}{\textsuperscript{$^1$}}
\newcommand{\leipzig}{\textsuperscript{$^2$}}
\newcommand{\martin}{\textsuperscript{$^{3}$}}

\author{
Matti Wiegmann\buw \qquad
Jennifer Rakete\leipzig \qquad
Magdalena Wolska\buw \qquad
{\bf Benno Stein}\buw \qquad
{\bf Martin Potthast}\martin \\[1ex]
\buw Bauhaus-Universit{\"a}t Weimar \quad
\leipzig Leipzig University \quad
\martin University of Kassel, hessian.AI, and ScaDS.AI
}

\date{}

\maketitle

\begin{abstract}
Trigger warnings are labels that preface documents with sensitive content if this content could be perceived as harmful by certain groups of readers. Since warnings about a document intuitively need to be shown before reading it, authors usually assign trigger warnings at the document level. What parts of their writing prompted them to assign a warning, however, remains unclear. We investigate for the first time the feasibility of identifying the triggering passages of a document, both manually and computationally. We create a dataset of 4,135~English passages, each annotated with one of eight common trigger warnings. In a large-scale evaluation, we then systematically evaluate the effectiveness of fine-tuned and few-shot classifiers, and their generalizability. We find that trigger annotation belongs to the group of subjective annotation tasks in NLP, and that automatic trigger classification remains challenging but feasible.


\warning{Warning. This paper shows example text relating to {\it Death}.}
\end{abstract}

\section{Introduction}

Online content is considered harmful if it has a negative emotional, psychological, or physical effect on people \cite{kirk:2022}. This applies regardless of whether the effect is intentional (as with hate speech) or unintentional. Since the publication of harmful content is in some cases justified or even important (as with news), affirmative action seems warranted. Several (online) communities have therefore developed a new type of affirmative action borrowed from trauma therapy: trigger warnings. Trigger warnings are freeform labels prefaced to a document by its author to indicate that it contains potentially harmful content. Commonly used warnings range from harmful concepts that can affect any individual, such as {\it Aggression}, {\it War}, or {\it Death}, to those that only affect certain groups of people, such as {\it Discrimination} and its sub-concepts {\it Misogyny}, {\it Racism}, or {\it Homophobia}. Given the availability of large-scale resources of author-labeled documents with triggering content, the task of automatic trigger warning assignment has recently been picked up (Section~\ref{related-work}).

Trigger warnings usually do not specify where the triggering content occurs in a document. This poses both theoretical and practical questions for the manual and automatic annotation or assignment of trigger warnings to text documents. For example, a concept such as {\it Death} may not be addressed throughout an entire document, but only occasionally. From an application perspective, document-level warnings prevent readers from reading the non-triggering parts of a document, which may undermine harm mitigation \cite{bridgland:2023}. From a linguistic perspective, triggers are at the pragmatic level of discourse and depends on context: Figure~\ref{figure-segments-with-death-warning} shows two examples addressing {\it Death}, their context influences annotator perception. Author-supplied ground truth, too, may hence comprise label noise, warranting closer inspection.

In this paper, we lay the foundation to investigate if and how trigger warnings can be reliably annotated and assigned at the passage level:
\Ni
We conduct a large annotation study to understand the challenges of trigger annotation, resulting in a dataset of 4,135~English five-sentence passages from the Webis Trigger Warning Corpus~2022 \cite{wiegmann:2023a}, annotated with eight triggers (Section~\ref{annotation}).
\Nii
We then systematically evaluate state-of-the-art classifiers in assigning various trigger warnings  and analyze their behavior regarding training data availability, label subjectivity, and generalization to unseen concepts (Section~\ref{classification}).%
\footnote{\raggedright All code and data are shared for reproducibility: \mbox{\href{https://github.com/MattiWe/passage-level-trigger-warnings}{github.com/MattiWe/passage-level-trigger-warnings}}}
We find that trigger warning annotation belongs to the subjective annotation tasks in %
NLP, and that classifying triggering passages requires a careful choice of the right model per warning (Section~\ref{results}). 

\section{Related Work}
\label{related-work}

Trigger warnings originate from trauma therapy~\cite{knox:2017} and have been studied in both clinical and educational contexts. A recent meta-study by \citet{bridgland:2023} provides an overview of the state of the art in these areas. The first computational trigger warning assignment approach \cite{stratta:2020} investigated an interaction design in a user study using a browser plugin (DeText) on generic websites, which was limited to a dictionary-based assignment of the {\it Sexual Assault} warning. Subsequently, \citet{wolska:22} examined the binary document classification of fan fiction documents using the trigger {\it Graphic Violence}. A taxonomy for multimedia triggers is the Narrative Experiences Online~(NEON) taxonomy proposed by \citet{charles:2022}, which contains 90~labels based on 136~guides from the web. \citet{wiegmann:2023a} later presented a 36-label taxonomy of trigger warnings, based on academic guidelines and a large-scale annotated corpus of fan fiction documents. Since the former does not provide annotated works, we draw on the latter's taxonomy and data. All these works consider the assignment of trigger warnings at the document level, while our focus is on the assignment of warnings at the passage level.

The classification of triggers also relates to that of other harmful content, such as toxic comments on Wikipedia articles \cite{wulczyn:2017, adams:2017}, verbal violence in YouTube and Reddit comments \cite{mollas:2020}, and the taxonomy of harmful online content \cite{banko:2020}, which overlaps with the above-mentioned taxonomies in the verbal categories. These works differ mainly in the delineation of harmfulness: who is affected (everyone, groups, or individuals), whether the harm is intentional or collateral, how the content is conveyed (in online speech such as microblogs or chats, or in long-form texts, such as comments, blogs, or narratives). The distinctive feature of trigger warnings is that they focus primarily on collateral harms to individuals, regardless of the form of transmission. \citet{kirk:2022} present a more comprehensive discussion of harmful (online) texts and their various aspects; we have adopted their recommendations for annotation.

\begin{figure}

\centering
\colorbox{blue!10}{%
\begin{minipage}{0.9\columnwidth}
The moans and cries of the damned assaulted one's ears. The stench of rotting and burning flesh assaulted one's nose. {\it The disfigurement of each hapless undead slave, some missing limbs, covered in blood and ooze, some naked, some with their skin missing, and more assaulted one's eyes.} Dark Magic, claws, and weapons assaulted one's body. And finally, the horror of it all assaulted one's mind.\\[0.5ex]
{\bf\small Annotation: positive instance}
\end{minipage}}

\bigskip
\colorbox{blue!10}{%
\begin{minipage}{0.9\columnwidth}
Zach would move onto whoever he met at the next frat party. His friends warned him beforehand that Jack is kinda weird. {\it Jack was in Corbyn's anatomy class and a cadaver has gone missing recently.} Of course, it was pure coincidence despite Corbyn insisting otherwise. It was just an unfortunate mishap, and it wouldn't deter Zach from his goal tonight. \\[0.5ex]
{\bf\small Annotation: negative instance}
\end{minipage}}

\caption{Two example passages from fan fiction stories annotated for the \warning{trigger warning {\it Death}}. The upper example was unanimously annotated as `positive', the lower as `negative'. The central sentence in italics was retrieved via keywords; preceding and following sentences serve as context. Figure~\ref{table-appendix-passage-examples} shows more examples.}\label{figure-segments-with-death-warning}

\end{figure}


As our research shows, another common feature of triggers and other harmful content is label subjectivity and annotator disagreement. In \citeauthor{sandri:2023}'s~\citeyear{sandri:2023} taxonomy of causes of disagreement in detecting offensive language use, `Sloppy Annotation' (which we mitigate by monitoring annotators and reducing task complexity to binary annotations) and `Missing Information' (which we mitigate by using passages instead of sentences) are mentioned as possible causes, but also the hard-to-avoid `Subjectivity' and `Ambiguity'. \cite{rottger:2022} note that there are several valid beliefs about labeling harmful content and that the ``descriptive'' annotation paradigm used in academic work (including ours) attempts to capture all of these beliefs, leading to disagreement among annotators. They suggest isolating a particular belief using a ``prescriptive'' annotation paradigm when desirable for subsequent application. Another approach is learning with disagreement, as studied in the  shared task by that name \cite{uma:2021} as well as that on `sexism recognition' \cite{plaza:2023}, captures multiple beliefs and models them explicitly. Similarly, \citet{davani:2022} study multitask learning over multiple annotators' votes to classify hate speech and emotions without loosing effectiveness.

\begin{table*}[t]%
\centering%
\small
\renewcommand{\tabcolsep}{1pt}%
\renewcommand{\arraystretch}{.9}%
\begin{tabular}{@{}l @{\hspace{-4pt}}cc c@{\hspace{-2pt}}c @{}}
\bf (a) \\
\toprule
\multicolumn{1}{@{}l@{}}{\bfseries Warning} & 
\multicolumn{2}{l}{\bfseries Passages} & 
\multicolumn{2}{@{}c@{}}{\bfseries Keyword}\\
\cmidrule(l{\tabcolsep}r{\tabcolsep}){2-3}\cmidrule(l{\tabcolsep}r{\tabcolsep}){4-5}
& 
num & len &
src & clean \\
\midrule
Violence        & 1,041    & 92    &\z29   & \z28   \\
Death           & \z\,544  & 83    &122    & \z50   \\
War             & \z\,827  & 95    &\z38   & \z25   \\
Abduction       & \z\,511  & 83    &\z20   & \z18   \\
\midrule
Racism          & \z\,267  & 90    &\z43   & \z37   \\
Homophobia      & \z\,313  & 79    &\z38   & \z27   \\
Misogyny        & \z\,377  & 84    &\z62   & \z55   \\
Ableism         & \z\,255  & 83    &\z47   & \z31   \\
\midrule
Total           & 4,135    & 88    &399    & 271  \\
\bottomrule
\end{tabular}%
\rule{0.4em}{0ex}
\begin{tabular}{@{}l @{\hspace{-4pt}}cc c@{\hspace{-0pt}}c @{\hspace{-2pt}}c@{\hspace{-2pt}}c @{}}
\bf (b) \\
\toprule
\multicolumn{1}{@{}l@{}}{\bfseries Warning} & 
\multicolumn{4}{@{}c@{}}{\bfseries Number of Positive Votes} &
\multicolumn{1}{l@{}}{\bfseries Time} & 
$\alpha$ \\
\cmidrule(l{\tabcolsep}r{\tabcolsep}){2-5}
&
0 & 1 & 2 & 3
\\
\midrule
Violence        & \z\,198(53\,\%)    & \z98(26\,\%)   & \z60(16\,\%)   & \z21(6\,\%)   & 39  &0.41   \\
Death           & \z\,\z82(31\,\%)   & \z60(22\,\%)   & \z37(14\,\%)   & \z88(33\,\%)  & 30  &0.36   \\
War             & \z\,\z86(34\,\%)   & \z82(32\,\%)   & \z53(21\,\%)   & \z34(13\,\%)  & 18  &0.22   \\
Abduction       & \z\,\z97(31\,\%)   & \z78(25\,\%)   & \z97(31\,\%)   & \z41(13\,\%)  & 29  &0.25   \\
\midrule
Racism          & \z\,310(56\,\%)    & 120(22\,\%)    & \z71(13\,\%)   & \z43(8\,\%)   & 29  &0.52   \\
Homophobia      & \z\,678(65\,\%)    & 175(17\,\%)    & 119(11\,\%)    & \z69(7\,\%)   & 27  &0.24   \\
Misogyny        & \z\,238(47\,\%)    & 141(28\,\%)    & \z94(18\,\%)   & \z38(7\,\%)   & 31  &0.25   \\
Ableism         & \z\,545(66\,\%)    & 180(22\,\%)    & \z83(10\,\%)   & \z19(2\,\%)   & 24  &0.25   \\
\midrule
Total           & 2,234(54\,\%)      & 935(23\,\%)    & 613(15\,\%)    & 353(9\,\%)    & 33  &0.35   \\
\bottomrule
\end{tabular}%
\rule{0.4em}{0ex}
\begin{tabular}{@{}l @{\hspace{-4pt}}cc @{}}
\bf (c)\\
\toprule
\multicolumn{1}{@{}l@{}}{\bfseries Warning} & 
\multicolumn{2}{@{}c@{}}{\bfseries Train} \\
\cmidrule(l{\tabcolsep}r{\tabcolsep}){2-3}
& ID & OOD \\
\midrule
Violence        & 1,001     & \z\,675   \\
Death           & \z\,504   & \z\,178   \\
War             & \z\,787   & \z\,202   \\
Abduction       & \z\,471   & \z\,252   \\
\midrule
Racism          & \z\,227   & \z\,\z68  \\
Homophobia      & \z\,273   & \z\,129   \\
Misogyny        & \z\,337   & \z\,152   \\
Ableism         & \z\,215   & \z\,162   \\
\midrule
Total           & 3,815     & 1,818   \\
\bottomrule
\end{tabular}%
\caption{Descriptive datatset statistics:
{\bf (a)}~Number and length (words) of the passages and the number of phrases in the source and cleaned keyword lists.
{\bf (b)}~Number and ratio of examples with $n$ positive votes, average annotation time per passage in seconds, and Krippendorff's $\alpha$ inter-annotator agreement. 
{\bf (c)}~Number of passages in the training and test splits for the in-distribution (ID) and out-of-distribution (OOD) settings.}%
\label{table-dataset-size-and-agreement}%
\end{table*} 

\section{Task Design}
\label{task-design}

Our approach to trigger annotation is the result of several small test runs and pilot studies that provided us with the insights to make design decisions for our annotation task. From these studies, we derived the following three key constraints for the task of trigger annotation at the passage level:  

\paragraph{Trigger Diversity}
The trigger warnings used in practice are based on the personal opinions and experiences of many authors and thus very diverse. The WTWC-22 organizes them into seven categories containing 36~warnings, encompassing hundreds of thousands of variants. Annotating a given passage with all 36~warnings, let alone all variants, has proven infeasible at scale under reasonable budget constraints. We therefore select eight common warnings for annotation, four each from the two most frequently assigned warning categories {\it Aggression} and {\it Discrimination} of the WTWC-22 taxonomy. From {\it Aggression} we use all four warnings {\it Death}, {\it Violence}, {\it Abduction} and {\it War}. From {\it Discrimination} we use the four most frequently assigned warnings {\it Misogyny}, {\it Racism}, {\it Homophobia} and {\it Ableism}. The former relate primarily to physical harm, the latter to psychological harm.

\begin{figure*}
\small
\centering
\colorbox{blue!10}{%
\rule{0.05\textwidth}{0ex}
\begin{minipage}{0.93\textwidth}
\begin{itemize}
    \setlength{\itemsep}{-0.2ex}
    \item[\bf Persona] You have suffered through {\it death} in the past and want to be warned before reading text that contains {\it death}.
    \item[\bf Definition] {\it Death includes graphic or implied descriptions of dead people or creatures, actions leading to death, experiences of dying, and descriptions of grief.} A Warning is required if the text evokes a negative experience through the actions or speech of characters, graphic descriptions, or through the narrative. A Warning is required if a vulnerable reader is harmed by reading the text.
    \item[\bf Demo.] Consider the following examples: \\
    Text: {\it He laid the cadaver back down, uselessly recalling every single thing he learned in first aid in an \dots} 
    \item[\bf Instruction] Assign `Warning: yes' if the following story piece needs a trigger warning for death, else assign `Warning: no'.
    \item[\bf Passage] Text: {\it The stench of rotting and burning flesh assaulted one's nose. The disfigurement of each \dots} 
\end{itemize}
\end{minipage}}

\caption{\warning{Warning Death.} Structure of both the annotation instructions and the prompts used in the few-shot experiments. Shown here is an example passage for the Death warning. The italics sections vary by instance.}\label{figure-instruction-and-prompt-parts}

\end{figure*}

\paragraph{Trigger Sparsity}
Harmful passages are often sparsely distributed over long documents,%
\footnote{E.g., many fan fiction works are as long as books.}
leading to class imbalance between positive an negative cases. We therefore resort to the commonly used approach of dictionary-based retrieval to obtain a sufficient number of positive candidate passages as detailed below.%
\footnote{Retrieval-guided annotation have previously been employed to create NLP resources.}
This entails selection bias, since very subtle cases of triggering content may not contain any explicit mention of one of the dictionary's phrases, and since some important phrases might be missing from our dictionaries, both leading to false negatives. We analyze this possibility with a targeted out-of-distribution experiment.

\paragraph{Trigger Severity}
Authors and readers are not equally sensitive to a potentially triggering concept (see Section~\ref{subjectivity}). A reader who enjoys horror may find fewer references to {\it Death} in fiction triggering than readers who do not. For instance, the positive example in Figure~\ref{figure-segments-with-death-warning} might be considered enticing rather than harmful. Therefore, when annotators are asked to make a descriptive binary decision, they often disagree. This is not uncommon in work on harmful content and, according to \citet{rottger:2022}, even desirable in an initial study like ours. However, this presents difficulties when assessing annotations and annotators, as inter-annotator agreement measures become less reliable due to different reader sensitivity to a particular trigger. This prevents the use of crowdsourcing platforms for annotation, as we cannot
\Ni
quantitatively evaluate annotators,
\Nii
train them reliably ahead of time, and
\Niii
provide appropriate support during annotation.
We therefore recruited local annotators and provided them with personal support.

\section{Dataset Construction}
\label{annotation}

We constructed a dataset of 4,135~passages with five consecutive sentences, each annotated with three binary human labels for one of eight selected trigger warnings. All passages originate from the Webis Trigger Warning Corpus~2022 (WTWC-22) \cite{wiegmann:2023a}, which compiles fan fiction documents with document-level trigger warnings. Figure~\ref{figure-segments-with-death-warning} shows example passages and Table~\ref{table-dataset-size-and-agreement} overviews our dataset.

\subsection{Retrieving Triggering Passage Candidates}
\label{collecting-triggering-segments}

We collected the passages for annotation using a keyword-based retrieval approach. We first constructed a keyword list for each of the eight considered warnings, then retrieved matching documents from WTWC-22 using a BM25-based initial retrieval%
\footnote{We retrieved the documents via elasticsearch 7.17.4 with the default BM25 scoring on the chapter text. We applied html-removal and stemming after the default pre-processing.} %
where the keywords served as query terms. From each document, we extracted the first sentence with a keyword match and added the two pre- and succeeding sentences as context. Adding context is often necessary understand the central sentence. We did not use the original (paragraph) segmentation of the documents, since they greatly varied in length, particularly since dialog turns are often written as single paragraph. The resulting five-sentence passages were de-duplicated across warnings and annotated in order of the BM25 score of the originating document. 

\paragraph{Keyword List Construction.}

We build a list of keywords and phrases for each of the eight trigger warnings by prompting {\tt gpt-3.5-turbo-0301} using the following prompt with slight morpho-syntactic adaptations for each warning: \\[0.5ex]
{\it Provide a list of <warning> language that must be avoided in a story for people that are triggered by <warning>. Return the list in JSON format.}

We cleaned this initial list manually to remove redundant (i.e. lexically similar) phrases, phrases that better match a different warning, or ambiguous phrases that produce many false positives, reducing the initial keywords by ca. 32\,\%. Table~\ref{table-dataset-size-and-agreement}a shows the number of words and phrases on each warning's list, before and after cleaning. We split each of the cleaned lists into two sets of keywords (in the middle according to the order returned by the model). The set identity is used to distinguish between in and out-of-distribution examples (cf. Section~\ref{experiment-datasets}). Table~\ref{table-appendix-keyword-bootstrapping-example} (\ref{appendix-dataset-eval}) illustrates the list construction process for the {\it Death} and {\it Misogyny} warnings.

\begin{table*}[h]%
\centering%
\small
\renewcommand{\tabcolsep}{4pt}%
\renewcommand{\arraystretch}{.9}%
\begin{tabular}{@{}l ccc ccc ccc ccc @{}}
\toprule
\multicolumn{1}{@{}l@{}}{\bfseries Warning} & 
\multicolumn{3}{@{}c@{}}{\bfseries Pairwise $\kappa$} & 
\multicolumn{3}{@{}c@{}}{\bfseries Pairwise Overlap} & 
\multicolumn{3}{@{}c@{}}{\bfseries Mean $\kappa$} & 
\multicolumn{3}{@{}c@{}}{\bfseries Positive Rate} 
\\
\cmidrule(l{\tabcolsep}r{\tabcolsep}){2-4}
\cmidrule(l{\tabcolsep}r{\tabcolsep}){5-7}
\cmidrule(l{\tabcolsep}r{\tabcolsep}){8-10}
\cmidrule(l{\tabcolsep}r{\tabcolsep}){11-13}
&
1-2 & 1-3 & 2-3 &
1-2 & 1-3 & 2-3 &
1 & 2 & 3 &
1 & 2 & 3
\\
\midrule
Violence     & 0.38 & 0.33 & 0.53    & 0.82 & 0.80 & 0.82        & 0.36 & 0.46 & 0.43    & 0.09 & 0.25 & 0.25    \\
Death        & 0.34 & 0.29 & 0.44    & 0.78 & 0.74 & 0.78        & 0.32 & 0.39 & 0.36    & 0.19 & 0.25 & 0.28    \\
War          & 0.18 & 0.08 & 0.38    & 0.78 & 0.77 & 0.81        & 0.13 & 0.28 & 0.23    & 0.12 & 0.19 & 0.18    \\
Abduction    & 0.27 & 0.31 & 0.26    & 0.77 & 0.67 & 0.65        & 0.29 & 0.26 & 0.29    & 0.20 & 0.20 & 0.47    \\
\midrule
Racism       & 0.66 & 0.43 & 0.46    & 0.83 & 0.71 & 0.73        & 0.55 & 0.56 & 0.44    & 0.44 & 0.49 & 0.56    \\
Homophobia   & 0.23 & 0.15 & 0.35    & 0.60 & 0.61 & 0.67        & 0.19 & 0.29 & 0.25    & 0.34 & 0.56 & 0.37    \\
Misogyny     & 0.39 & 0.14 & 0.25    & 0.74 & 0.69 & 0.73        & 0.26 & 0.32 & 0.20    & 0.31 & 0.30 & 0.14    \\
Ableism      & 0.13 & 0.31 & 0.35    & 0.57 & 0.66 & 0.71        & 0.22 & 0.24 & 0.33    & 0.49 & 0.33 & 0.31    \\
\midrule
Total        & 0.36 & 0.28 & 0.41    & 0.76 & 0.73 & 0.76        & 0.32 & 0.38 & 0.34    & 0.21 & 0.28 & 0.29    \\
\bottomrule
\end{tabular}%
\caption{Additional measures of annotation quality. Shown are the pairwise Cohen's $\kappa$ between the annotators (label wise, since an annotator always rated one label completely), the fraction of overlapping annotations between two annotators, the mean Cohen's $\kappa$ agreement of each annotator with the other two, and each annotator's rate of positive annotations. Note that the annotators differ across warnings.}%
\label{table-appendix-extended-agreement}%
\end{table*} 

\subsection{Passage Annotation}
\label{annotating-the-segments}

Figure~\ref{figure-segments-with-death-warning} shows two passages, one for each annotation decision. The collected passages were annotated as either positive (requires a warning) or negative (does not require a warning) by three different annotators each. We decided on the final annotation based on the majority of the votes. 

Instead of annotating a fixed number of passages for each warning, the first annotator of each warning rated continuously, in order of the BM25 score of the source document, until 50 passages were marked positive in each set (100 per warning). All passages marked either positive or negative by the first annotator were then also rated by the other annotators. This step is necessary because the ratio of positive-to-negative passages varies between the warnings and is very low (1:10) for some warnings like {\it Death} and {\it War} in set 2, so having an equal number of passages for each warning would results either in a very high number of annotations or a low number of positive examples for some classes.

\paragraph{Annotation Task Design}

The annotation task was designed as a binary classification decision.%
\footnote{We used Label Studio 1.6.0rc5 as the annotation system.} %
Annotators were presented with one passage at a time, an example of which is shown in Figure~\ref{figure-instruction-and-prompt-parts}. The markable consisted of five parts:
\Ni
a description of the `Persona' which the annotators should adapt for their decision, 
\Nii
the `Definition' of the warning which we created manually to have a similar length and scope,
\Niii
two `Demonstrations', one positive and one negative, selected by the authors,
\Niv
the `Instructions' explained the binary classification decision, and
\Nv
the `Passage' to be annotated.

\paragraph{Annotator Instruction and Monitoring}

In total, we recruited seven permanent annotators of different genders and backgrounds. The passages were assigned based on the warning, i.e. annotator 1 rated all {\it Misogyny} passages, annotator 2 all {\it Racism} passages, and so on. The annotators were allowed to freely choose the warning they wanted to annotate, however, they were also asked to choose a warning that they were familiar with, if possible.

Following \citet{kirk:2022}, we tried to reduce the stress on the annotators in two ways: First, we set no deadlines and we encouraged annotators to work in small batches over a longer period (6 weeks). Second, we arranged for weekly personal meetings with the annotators to discuss unclear cases, to monitor their well-being regarding the task, and to refine the shared understanding of the task through open discussion. The annotators were allowed to re-iterate and modify their annotations at any time.

\subsection{Evaluation}

We evaluate 
\Ni
if the keyword-based passage retrieval is effective,
\Nii
if the dataset is large and balanced enough for machine learning, and 
\Niii
if the task design facilitates high-quality annotations.

\paragraph{Passage Retrieval}

The annotation results in Table~\ref{table-dataset-size-and-agreement}b and Table~\ref{table-appendix-extended-agreement} show that~1,901 (46\,\%) of the retrieved passages received at least one positive vote, so a sensitive reader might desire a warning. The highest positive rate~(PR) is~69\,\% for {\it Homophobia} and {\it Racism}, the lowest is~34\,\% for {\it War}. Considering that most passages in any given document are negative, we consider our keyword-based passage retrieval effective in recalling good annotation candidates. However, some ambiguous keywords like `hit' ({\it Violence}) or `occupy' ({\it War}) retrieve many off-topic examples, lowering precision. Although these are often easy to annotate, better filtering would reduce annotation costs.

\paragraph{Size}

Table~\ref{table-dataset-size-and-agreement}b shows that negative instances are more frequent than positives. {\it Discrimination} warnings have a higher positive ratio than {\it Aggression} warnings. While the complete dataset is nearly balanced (45\,\% PR) under a 1-vote threshold, the data skew towards negatives (23\,\% PR) under a 2-vote threshold. The balance varies between keywords, which is a problem for standard splits because the test or validation sets may have few positive (Set~2 {\it War}) or negative (Set~1 {\it Death}) instances. Instead, we opted for cross-validation.

Causes for the imbalance are, first, some keywords retrieve more severe passages. For example, {\it Racism} Set~1 contains many strong slurs and has a high positive rate (99\,\%), while {\it Death} Set~2 (31\,\% positive rate) contains many fantasy concepts that are also used in harmless contexts. Second, annotators differ in sensitivity to some warnings (see Table~\ref{table-appendix-extended-agreement}), like Rater~1 on {\it Violence} (9\,\% PR compared to 25\,\%) or Rater~3 on {\it Abduction} (47\,\% to~20\,\%).

\paragraph{Annotation Quality}\label{annotation-qualty}

Table~\ref{table-dataset-size-and-agreement}b and Table~\ref{table-appendix-extended-agreement} show a chance-corrected inter-annotator agreement (0.22--0.52, mean 0.35 Krippendorff's~$\alpha$). This indicates a ``fair agreement'', which is consistent with similar tasks (0.34--0.58 Cohen's~$\kappa$ over two annotators for binary offensive language by \citet{pitenis:2020}; 0.20 Fleiss'~$\kappa$ over 20~annotators for binary hateful language by \citet{rottger:2022}). We expected a reasonable degree of disagreement due to the subjective nature of trigger warnings. Text appears harmful to individuals based on their personal lived experience, which, naturally, varies between annotators. We still evaluated our annotators quantitatively by measuring the pairwise overlap (0.74--0.77 mean across all warnings) and the pairwise agreement (0.31--0.44 Cohen's~$\kappa$). One annotator systematically disagreed with the other two ({\it Racism}, Annotator~3 with mean $\kappa$ of~-0.13), so that the annotations were repeated by another.

\subsection{Subjectivity and Annotator Beliefs}\label{subjectivity}

\begin{figure*}[t]
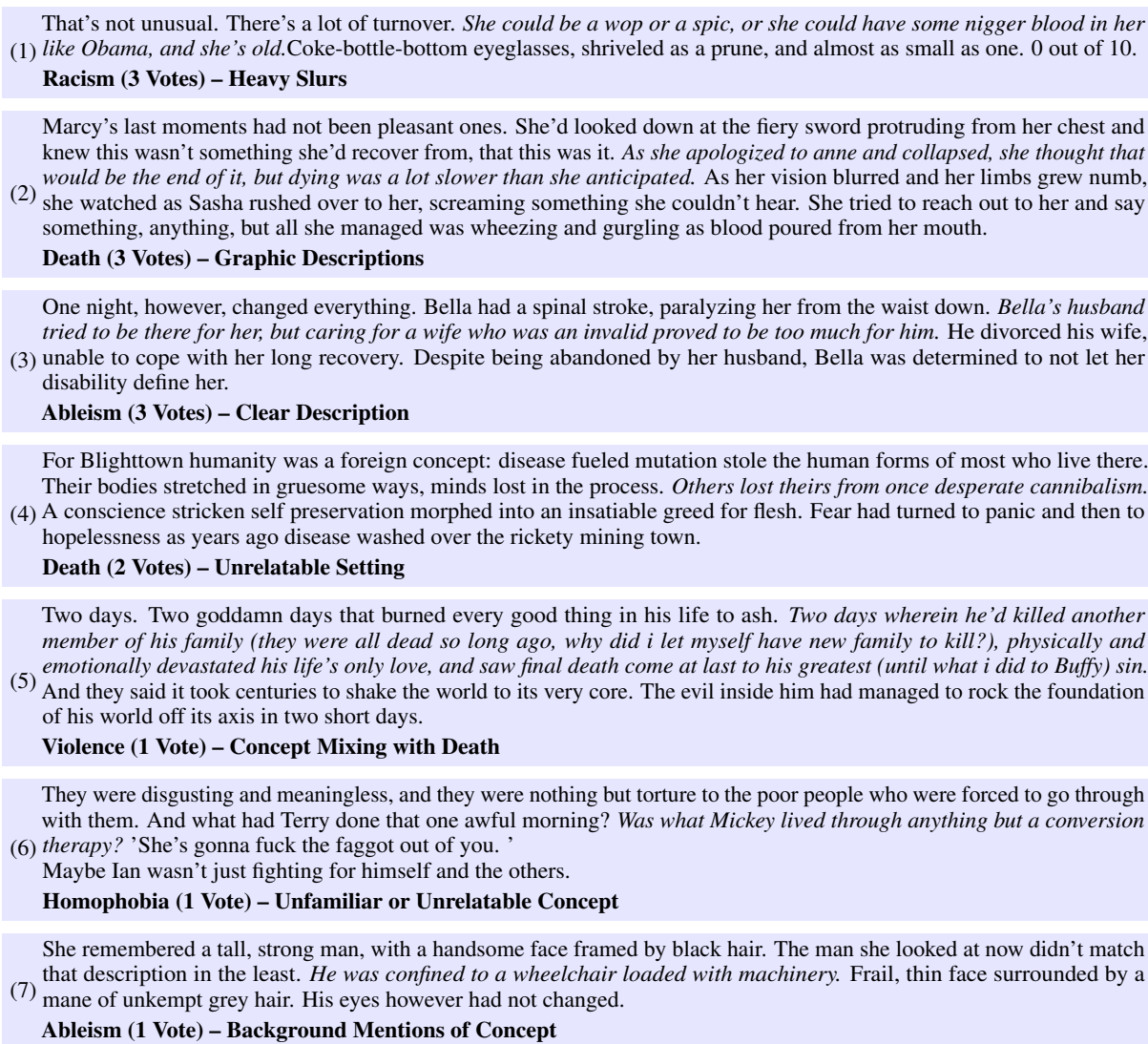
%
\small
\centering
\colorbox{blue!10}{%
(1) \begin{minipage}{0.95\textwidth}
That's not unusual. There's a lot of turnover. {\it She could be a wop or a spic, or she could have some nigger blood in her like Obama, and she's old.}Coke-bottle-bottom eyeglasses, shriveled as a prune, and almost as small as one. 0 out of 10. \\[0.5ex]
{\bf\small Racism (3 Votes) -- Heavy Slurs}
\end{minipage}}
\\[1ex]
%
\colorbox{blue!10}{%
(2) \begin{minipage}{0.95\textwidth}
Marcy's last moments had not been pleasant ones. She'd looked down at the fiery sword protruding from her chest and knew this wasn't something she'd recover from, that this was it. {\it As she apologized to anne and collapsed, she thought that would be the end of it, but dying was a lot slower than she anticipated.} As her vision blurred and her limbs grew numb, she watched as Sasha rushed over to her, screaming something she couldn't hear. She tried to reach out to her and say something, anything, but all she managed was wheezing and gurgling as blood poured from her mouth. \\[0.5ex]
{\bf\small Death (3 Votes) -- Graphic Descriptions}
\end{minipage}}
\\[1ex]
\colorbox{blue!10}{%
(3) \begin{minipage}{0.95\textwidth}
One night, however, changed everything. Bella had a spinal stroke, paralyzing her from the waist down. {\it Bella's husband tried to be there for her, but caring for a wife who was an invalid proved to be too much for him.} He divorced his wife, unable to cope with her long recovery. Despite being abandoned by her husband, Bella was determined to not let her disability define her. \\[0.5ex]
{\bf\small Ableism (3 Votes) -- Clear Description}
\end{minipage}}
\\[1ex]
\colorbox{blue!10}{%
(4) \begin{minipage}{0.95\textwidth}
For Blighttown humanity was a foreign concept: disease fueled mutation stole the human forms of most who live there. Their bodies stretched in gruesome ways, minds lost in the process. {\it Others lost theirs from once desperate cannibalism.} A conscience stricken self preservation morphed into an insatiable greed for flesh. Fear had turned to panic and then to hopelessness as years ago disease washed over the rickety mining town.
\\[0.5ex]
{\bf\small Death (2 Votes) -- Unrelatable Setting}
\end{minipage}}
\\[1ex]
\colorbox{blue!10}{%
(5) \begin{minipage}{0.95\textwidth}
Two days. Two goddamn days that burned every good thing in his life to ash. {\it Two days wherein he'd killed another member of his family (they were all dead so long ago, why did i let myself have new family to kill?), physically and emotionally devastated his life's only love, and saw final death come at last to his greatest (until what i did to Buffy) sin.} And they said it took centuries to shake the world to its very core. The evil inside him had managed to rock the foundation of his world off its axis in two short days. 
\\[0.5ex]
{\bf\small Violence (1 Vote) -- Concept Mixing with Death}
\end{minipage}}
\\[1ex]
\colorbox{blue!10}{%
(6) \begin{minipage}{0.95\textwidth}
They were disgusting and meaningless, and they were nothing but torture to the poor people who were forced to go through with them. And what had Terry done that one awful morning? {\it Was what Mickey lived through anything but a conversion therapy?} 'She's gonna fuck the faggot out of you. ' \\
Maybe Ian wasn't just fighting for himself and the others.
\\[0.5ex]
{\bf\small Homophobia (1 Vote) -- Unfamiliar or Unrelatable Concept}
\end{minipage}}
\\[1ex]
\colorbox{blue!10}{%
(7) \begin{minipage}{0.95\textwidth}
She remembered a tall, strong man, with a handsome face framed by black hair. The man she looked at now didn't match that description in the least. {\it He was confined to a wheelchair loaded with machinery. }Frail, thin face surrounded by a mane of unkempt grey hair. His eyes however had not changed.
\\[0.5ex]
{\bf\small Ableism (1 Vote) -- Background Mentions of Concept}
\end{minipage}}

\caption{\warning{Trigger Warning.} Selected example passages with assigned warning and number of positive votes. The center sentence that was retrieved via keywords is is italics.}%
\label{table-appendix-passage-examples}%
\end{figure*} 

Table~\ref{table-dataset-size-and-agreement}b shows the passage count by number of positive votes. Out of all passages, 55\,\%~were unanimously negative, but only~9\,\% unanimously positive. The negatives are partially explained by off-topic passages retrieved by ambiguous keywords, although this does not account for all negatives (see Figure~\ref{table-appendix-error-cases-examples}). The variance across positive votes is explained by varying annotator sensitivity and a different belief of when a warning is required. Figure~\ref{table-appendix-passage-examples} shows typical positive passages. Unanimous positive passages are often severe cases with heavy use of slurs and very graphic descriptions. Non-unanimous examples feature unfamiliar or unrelatable settings (fantasy, science fiction) or concepts (`conversion therapy'), background or implied mentions, or a mix between warnings which confuse the annotators. Contributing to this are keywords like `undead', `reanimation', or `necromancy' often contained in fantastical scenes (see Table~\ref{table-appendix-keyword-bootstrapping-example}). Removing those keywords would reduce annotation load and annotator uncertainty but would also reduce recall, remove positive passages, and further limit the scope of the concepts for model training.

The vote aggregation threshold impacts the characteristics of a classifier. A low threshold (1~vote) produces recall-oriented classifiers for warning assignment, while a high threshold (2--3~votes) produces precision-oriented classifiers suitable for moderation tools. We assume that resolving annotator disagreement will require either a form of personalization with a per-warning known sensitivity, or a prescriptive transformation of the annotation task, where the triggers are intensionally defined. 

\section{Passage Classification}\label{classification}

We structure our experiments around four relevant design decisions and analyze their effect on the performance across models and warnings.

\paragraph{Class Modeling} We comparatively evaluate binary, multiclass, and multilabel modeling via fine-tuned classification. We formulated the annotation task as binary classification: given a passage and a warning, decide if the warning should be assigned. Hence, our reference baseline and our evaluation setting is binary classification. However, binary classification is expensive to scale since each new warning needs a new classifier, and new training data, and each classifier does not use most of the annotated data. Under multiclass modeling, only one classifier is needed so the training data is better utilized, although such a classifier can only predict one warning per passage. Multilabel modeling combines these advantages of binary and multiclass, but is more difficult to train and our (binary) data might not be sufficient.  

\paragraph{Fine-tuning vs. Few-shot Learning} We comparatively evaluate six few-shot prompted generative LLMs which can scale to a high-dimensional warning taxonomy like NEON. Trigger warnings are potentially open-class and might require personalization, which is difficult to scale to with model fine-tuning. 

\paragraph{Vote Aggregation} We comparatively evaluate majority (2 votes count as positive) and minority voting (1 vote counts as positive) across all other experimental configurations. Our dataset evaluation suggested that examples containing clear and intense harm often receive multiple positive votes (24\% with 2--3), while examples with one vote (23\%) often contain only mild or implicit harm. The minimum number of votes required for an example to count as positive for classification will thus influence the sensitivity of the classifier.

\paragraph{Keyword Distribution} We evaluate all experiments across in-distribution (ID) and out-of-distribution (OOD) samples of the dataset. Using keyword-based filtering to pre-select passages may limit how well the classifiers generalize to unseen examples that match the respective warnings but not the concepts captured by the keywords. We simulate this situation by splitting the keywords (and the passages retrieved using them) into two non-overlapping sets (cf.~\ref{collecting-triggering-segments}) and sampling datasets once from both sets combined and once separately. 

\subsection{Experiment Datasets}\label{experiment-datasets}

We compiled a total of 16 initial, unbalanced datasets, one for each of the eight warnings with majority and minority vote aggregation respectively. To get comparable results, we sampled balanced test sets with 20 positive and 20 negative instances for each of the 16 unbalanced datasets, which is the limit imposed by the smallest set (Misogyny Set 2). Since 40 instances are too few to get stable results, we created six random folds for a 5-fold Monte Carlo cross-validation with a sixth fold for parameter tuning. 

For the in-distribution experiments, the test data are randomly drawn from all examples (i.e. they share passages retrieved by both keyword sets), while all other instances remain for training. In the out-of-distribution experiments, the test data are randomly drawn from all passages retrieved by keyword set 2, all other set 2 instances are discarded and all set 1 instances remain for training. Table~\ref{table-dataset-size-and-agreement}c shows the number of examples in the training and test datasets.

\subsection{Models and Training}

We evaluate five fine-tuning-based classification models {\tt Binary}, {\tt Binary+} (which includes negative instances from other classes), {\tt Multiclass}, {\tt Multilabel} (without any all-negative instances) and {\tt Multilabel+} (with all-negative instances) and six generative LLMs {\tt GPT\,3.5}, {\tt GPT\,4}, {\tt Mistral\,7B}, {\tt Mixtral\,7x8B}, {\tt Llama\,7B}, and {\tt Llama\,13B}. We implemented the models using Huggingface's {\it transformers} library (4.35.0). We prompted GPT via OpenAI's API. 

\paragraph{\tt Binary}

We trained eight binary classification models for each configuration, one for each warning, on all positive and negative warning-passage pairs of the respective warning. We also trained a version {\tt Binary+} where each classifier was trained on the positive instances of the respective warning and the negative instances of all other warnings. We hypothesize that this expansion might improve the classification in the out-of-distribution configurations.

\paragraph{\tt Multiclass}

We trained one multiclass classifier by combining the eight training datasets: We assigned each positive instance a class label for it's warning (class 0--7) and all negative instances to class 8. We test the models on the binary datasets. We count a prediction as positive if it predicts the class corresponding to the dataset's warning and all other classes as negative (class 8). For example, when predicting multiclass for {\it Death}, then all predicted classes except {\it Death} count as negative predictions. This is a limitation imposed by annotating the dataset in a binary fashion.  

\paragraph{\tt Multilabel}

We trained one multilabel classifier by combining the eight training datasets and converting the positive binary class labels to respective one-hot label vectors and discarding all negative labels. This corresponds to a one-class paradigm which can be easily expanded for new warnings. We also trained a version {\tt Multilabel+} which includes all negative examples with a zero vector as label (see limitations below). We test the models, similarly to multiclass, by only considering the models prediction for a class if the instance was annotated for that class, i.e. we ignore all predictions for other classes. 

Since the training passages are only annotated in a binary way for the respective warning and there is no overlap between passages across warnings, converting the examples to multilabel class vectors introduces errors: If a passage would be positive for a warning it was not annotated for, the class vector becomes incomplete and might confuse the classifier. Although the multilabel class modeling is the most practically convenient, it is also not very promising to reach a high performance due to these issues. Adding negative examples for the {\tt multilabel+} classifier will likely exacerbate this problem (cf. Figure~\ref{table-appendix-passage-examples}(4)). 

However, since the passages are short, they are usually only positive for multiple warnings from within the same category (either {\it Aggression} or {\it Discrimination}). An example selected for {\tt Death} may be positive for {\it War} or {\it Violence}, but is likely not positive for {\it Racism}. That means we can mitigate adding many false positives by splitting the {\tt multilabel+} classifier into two: one multilabel classifier for all {\it Aggression} warnings where only negatives from {\it Discrimination} are added, and vice versa.

\paragraph{\tt GPT 3.5 and GPT 4}

We prompted the models `gpt-3.5-turbo-0125' and `gpt-4-0125-preview' using the OpenAI API as described in Section~\ref{prompt-ablation}. We only prompted the models for the respective test splits across all warnings and folds.

\paragraph{\tt Mistral 7B and Mixtral 7x8B} 

We implemented both Mistral models via Huggingface's {\it transformers} library, using the `mistralai/Mistral-7B-Instruct-v0.2' and `mistralai/Mixtral-8x7B-Instruct-v0.1' checkpoints respectively. We encoded the prompt using Mistral's chat template (without system message) and prefixed the prompt with: {\it You are a classification model that only answers with 'yes' or 'no'}. {\tt Mistral 7B} was tested on one A100 40GB. {\tt Mixtral 7x8B} was loaded with 8-bit quantization and tested on three A100 40GB.

\paragraph{\tt Llama 7B and 13B}

We implemented both Llama models via Huggingface's {\it transformers} library, using the `meta-llama/Llama-2-7b-chat-hf' and `meta-llama/Llama-2-13b-chat-hf' checkpoints respectively. We used the prompt directly and without the chat template, because the models produces difficult-to-parse output when using the chat template as we did with Mistral. {\tt Llama 7B} was tested on two A100 40GB with a batch size of 12. {\tt Llama 12B} was loaded using the A100 GPUs and batch sizes of 8.

\begin{figure*}[t]
    \includegraphics[width=\textwidth]{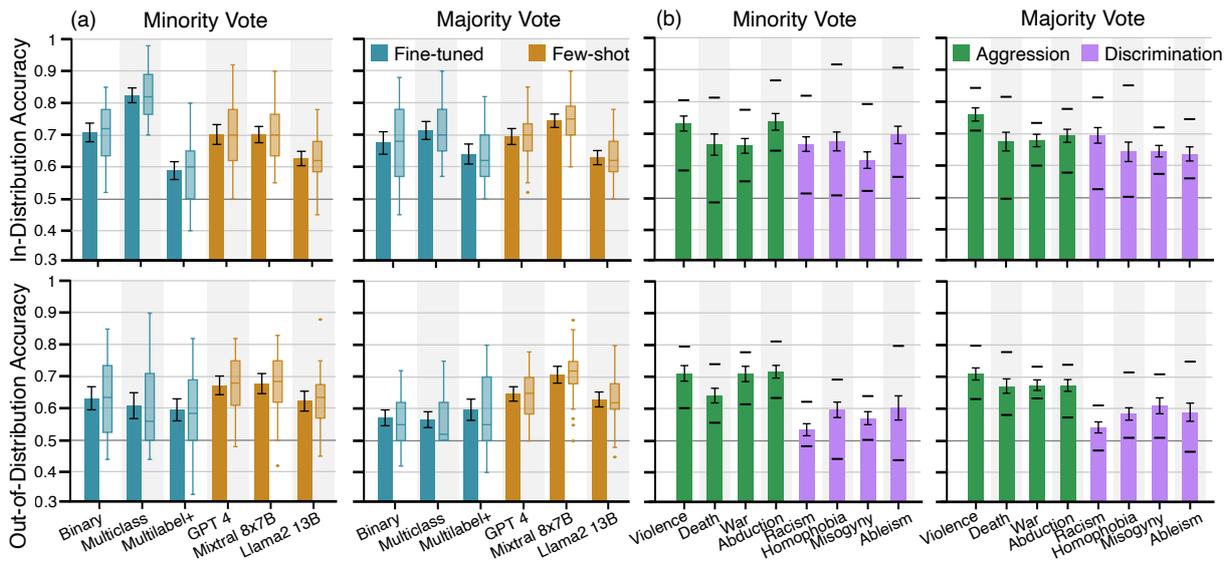}
    \caption{Overview of the experimental results across distribution (in vs. out-of) and vote aggregation (minority vs. majority). {\bf (a)} The bar charts show the mean accuracy for the models across all warnings with 95\% t-estimated confidence interval. The box plots show the quartiles and outliers of all folds and warnings. {\bf (b)} The bar chars show the mean accuracy for the warnings across all models with 95\% t-estimated confidence interval. The upper and lower ticks show the best and worst model's performance. Table~\ref{table-appendix-classification-results} (Appendix~\ref{appendix-a}) shows the full results.}\label{prediction-performance-across-models}  
\end{figure*}

\subsection{Evaluation}

We test all models on the balanced, binary test datasets for ease of comparison and report accuracy and positive rate. For multiclass and multilabel models we only count the predictions relating to the respective warning (cf~\ref{model-implementation}). We report the mean performance across all five folds for each model and macro-averages across models and warnings.

\section{Results and Discussion}
\label{results}

Figure~\ref{prediction-performance-across-models} and Table~\ref{table-appendix-classification-results} (\ref{appendix-classification-eval}) show that models typically score between ca. 0.6--0.7 accuracy on average across all warnings, depending on the model and data configuration, while the top models per warning score between ca. 0.7--0.8. The highest-scoring models are {\tt Multiclass} on in-distribution data (ID) with minority voting (MinV) (0.82), {\tt Mixtral} on ID (0.74), and out-of-distribution (OD) (0.71) majority voting (MajV), and {\tt GPT\.3.5} with 0.7 on OD/MinV. The lowest mean scores are barely above random and occur for some worst-case configurations. The best per-warning score is 0.91 ({\tt Multiclass} on {\it Homophobia}). {\tt Multiclass} is the most effective model in 10 of 32 cases across warnings and configurations, followed by {\tt Mixtral} with 9. The scores vary between warnings, where {\it Violence} has the highest (0.70--0.72 mean accuracy) and {\it Racism} the lowest scores (0.52--0.69). {\it Aggression} warnings score higher than {\it Discrimination}. 

\begin{figure*}[t]%
\small
\centering
\colorbox{blue!10}{%
(1) \begin{minipage}{0.95\textwidth}
It's easier for him to carry what he needs to avoid suspicions. Kunimi is required to pay for his medical school tuition tomorrow. Which is a cadaver. Kunimi doesn't understand what's the point of finding dead bodies anymore. His classmates practically dug all the bodies from all the cemeteries, but it's still not enough.
\\[0.5ex]
{\bf\small Death (0 Votes) -- Not Perceived as Harmful}
\end{minipage}}
\\[1ex]
\colorbox{blue!10}{%
(2) \begin{minipage}{0.95\textwidth}
Many, many things did happen to me. the hardest part is simply in starting, because once you do, the words come flying out. I asked myself for a long time whether i would be able to return to a civilian life. How could i get up in the morning if i didn't have a purpose, if i wasn't useful? I'd been a soldier, and a doctor, and a bullet through my shoulder ripped me of both titles.
\\[0.5ex]
{\bf\small War (0 Votes) -- Implicit or Background Mentions}
\end{minipage}}
\\[1ex]
\colorbox{blue!10}{%
(3) \begin{minipage}{0.95\textwidth}
The military bases also held the space force academy, a supposedly nonpartisan association that was clearly held and run by the allies of the United States. If Luna decided to put sanctions on Chinese trade, that also included chinese traders. Directly, that would affect thousands of people, but with sanctions came racism. The workers that lived on Luna, Mars, Deimos, Phobos, and the other new colonies would be ostracized, meaning millions would be left in the cold. Luna would be cutting off all shipments to and from China, which meant China wouldn't be trading with any colonies in the solar system.
\\[0.5ex]
{\bf\small Racism (0 Votes) -- Implicit or Background Mentions}
\end{minipage}}
\\[1ex]
\colorbox{blue!10}{%
(4) \begin{minipage}{0.95\textwidth}
'I remember my last holder complaining about that too...' 
'It sucks and it hurts and there's no point! and i'd go to the doctor and get that pill thing to stop it, but my dad won't let me, and... and speaking of my dad, you heard what he was saying earlier...' Fluff nodded again. 
Sitting in Alix's pocket all day, there was no way she couldn't have heard it - the usual why can't you be more ladylike? You're not a little child anymore, you need to stop being so immature! Can't you be more like your friends?
\\[0.5ex]
{\bf\small Misogyny (1 Vote) -- Edge Cases with Disagreement}
\end{minipage}}
\\[1ex]
\colorbox{blue!10}{%
(5) \begin{minipage}{0.95\textwidth}
Usually, they lasted longer than that, a month, at most. Whenever she started to trust them, they would disappear. Some found murdered, others never seen again. Ramsay liked to play with her, giving and taking anything she found comfort in. There was Theon, but she found no comfort in him, only pity.
\\[0.5ex]
{\bf\small Violence (1 Vote) -- Edge Cases with Disagreement}
\end{minipage}}
\\[1ex]
\caption{\warning{Trigger Warning.} Selected example passages that were always misclassified by all models.}%
\label{table-appendix-error-cases-examples}%
\end{figure*} 
\paragraph{Class Modeling}
Model efficiency varies significantly between modeling strategies: {\tt Multiclass} outperforms {\tt Binary} (0.11 pp. for ID/MinV;) and {\tt Multilabel+} (0.23 pp. for ID/MinV; 0.07 pp. ID/MajV) and {\tt Binary} outperforms {\tt Multilabel} (0.09 for OOD/MinV). All other differences are not significant. Adding more negative examples to {\tt Binary} shows no significant difference and adding negative examples to {\tt Multilabel} improves the measured performance within the 95\% CI by 0.01--0.06. Although {\tt Multiclass} is the most effective fine-tuned model, it works only if the warnings are separable and rarely overlap. Since this is often not the case (cf. Figure~\ref{table-appendix-passage-examples} (5)), the binary model is a more realistic alternative.

\paragraph{Fine-tuning vs. Few-shot Learning}

The results show that all fine-tuned models perform equal or worse than the best few-shot models ({\tt Mixtral}) in the OOD configuration. In the ID configuration, the fine-tuned models score comparative or higher ({\tt Multilabel} for ID/MinV). The fine-tuned models are also more efficient in time and energy and should be preferred whenever possible. 

Another notable difference is that all few-shot models have a higher average positive rate (PR) than the fine-tuned model (9--49 pp.), so the generative LLMs can be considered more sensitive. The reason for this is not clear, although explicit harm avoidance during training is a possible explanation.  

\paragraph{Vote Aggregation} 

Increasing the aggregation threshold from one to two significantly increases the accuracy for {\tt Multilabel+} and decreases the accuracy for {\tt Multiclass}, for {\it Ableism}, and for {\it Abduction} (all ID). %
In addition, the inter-quartile range of the folds (box plots in Figure~\ref{prediction-performance-across-models}) is smaller for MajV. This is likely explained by the reduced number of positive examples, which reduced the variance of the examples in the folds since there are much fewer positive examples to draw from.

Increasing the threshold also reduces the PR (3--28 pp.) across all fine-tuned models and increases the PR (6--10 pp.) across all few-shot models. The increase is likely explained by the overall increases of instances with 1+ votes increases with higher thresholds, since these get samples as the negative class and, since the the prompt did not change, this increases the few-shot models' positive rates. 

\paragraph{Keyword Distribution}

Overall, all models are less effective in the OOD configuration. This is expected behavior for fine-tuned models, since they can rely less on learned lexical cues, but not necessarily for few-shot models. A possible explanation is that we split the keywords in the order of generation, so keywords that are strongly associated with a trigger are in set 1. The OOD test set then contains passages that are less strongly associated with the trigger, thus reducing the few-shot scores.

In the OOD configurations, all fine-tuned models are significantly worse by ca. 0.1--0.2 pp. on average with an increased spread of per-warning scores while the few-shot models score as on ID (except for a reduction on {\it Racism}). Fine-tuned models are still competitive for most {\it Aggression} but ineffective for {\it Discrimination} warnings with barely above random accuracy and very low PRs. This is likely due to systematic differences between the keywords (cf. \ref{table-appendix-keyword-bootstrapping-example}). While the {\it Aggression} keywords are somewhat consistent between sets, {\it Discrimination} keywords often start with a series of slurs (Set 1) and become more conceptual later (Set 2), leading to a topical gap between OOD training and test. This was the goal of the configuration and shows that fine-tuned models generalize poorly in this case, highlighting the importance of diverse training data.

\subsection{Causes of Misclassification}

Table~\ref{table-appendix-error-cases-examples} shows selected passages that are always misclassified. Cases of systematic misclassification contain passages that %
\Ni are on-topic but are never rated as triggering by the annotators, %
\Nii mention the topic implicitly or in the background (like historic discrimination, descriptions of the setting) but are never rated as triggering by the annotators, and %
\Niii with edge cases which are rated positively by only one annotator. %
Misclassified instances seem to align with annotator disagreement. Across all few-shot models, many (56\%/75\%) instances with 1 or 3 votes are always classified correctly, but only 15\% with 1 vote. {\tt Mixtral} classifies 70--91\% of passages with 1/2/3-votes correctly, but only 30\% of 1-vote instances. The fine-tuned models, too, more often misclassify examples with disagreement but less extreme: {\tt Binary} (ID) is correct for 66\% of 3-vote and only 44\% on 2-vote instances. 

\section{Conclusion}\label{conclusion}

In this paper, we seek to identify the exact text passages that prompt authors to preface their works with trigger warnings. We model the task as binary classification and create a dataset of 4,135~English passages, each annotated by three human votes across 8 trigger warnings. We investigate how different beliefs about triggers contribute to the assignment of warnings by quantitatively and qualitatively analyzing annotator disagreement and the behavior of 11 classifiers across different design decisions: Vote Aggregation, Keyword Distribution, Class Modeling, and Training vs. Prompting.

Our keyword-based passage retrieval identifies many positive instances, from severe and graphic passages with high agreement to mild and implicit ones where the annotators' opinions differ based on their personal sensitivities and beliefs about the need for trigger warnings. We find that where annotators disagree, classification errors occur more frequently. Furthermore, we can experimentally show that, first, diverse training data is required for models to generalize well to unseen concepts and rare triggers, second, few-shot models like Mixtral are competitive (albeit computationally expensive), especially for unseen triggers, and, third, fine-tuned models are often still the best models for individual warnings and in certain configurations. It is advantageous to select a model specifically for the targeted warning or configuration, depending on the goal, like deleting content vs. adding warnings, or on the targeted sensitivity. 

\enlargethispage{\baselineskip}
In conclusion, we question whether authors can determine a trigger warning equally well for every reader. Future work may include corresponding investigations. Personalizing trigger warning assignment seems a fruitful computational direction.

\section*{Limitations}

We consider four limitations.
First, we only consider eight warnings from two categories in this study, all of them from more frequent cases. The WTWC-22 corpus contains 36 warnings from seven categories, while other taxonomies list up to 90. Consequentially, our strategy to generate keywords and collect passages to annotate might not work for rare or very individual warnings. Besides, the annotation instructions and prompts may also not generalize to some of the warnings, and model performance may differ. 

Second, we recruited annotators from a population of computer science students which is a shared demographic and hence a source of bias. All annotations were done by individuals who are not (clinically) affected by any trigger but who have different sensitivities to the different triggering concepts. The latter limitation holds for almost all trigger warnings which are usually assigned by the authors of the content. However, more reliable annotations would require that annotators are actually triggered by the concepts described by the annotation categories. We know of no practical and ethically feasible way to create such annotations.

Third, we cannot make any claims about which parts of the annotation instructions and the prompts (cf. Figure~\ref{figure-instruction-and-prompt-parts}) influence the performance of the annotators or the models beyond our ablation study.

Fourth, we did not quantitatively evaluate if our annotation instructions are the most optimal or if a different annotation paradigm would reduce the subjectivity of the task without limiting the captured beliefs about trigger warnings. 

\section*{Impact Statement}

As with any work on harmful content in the age of generative AI, our data, code, and insights may used in bad faith to generate harmful content or for adversarial engineering to trick detection algorithms. The artifacts we use (WTWC-22) are for academic use only, as are the artifacts we created: our data and code. 

It is safe to assume that due to the inherently subjective nature of the phenomenon of triggering, future work will focus on personalized decision-making for harmful content classification.
However, personalization here
is a highly sensitive issue since it requires deep knowledge about the individuals and their vulnerabilities.

We followed best practices when annotating harmful content for this work. The student annotators were not paid, besides teaching credit of which annotations were only part. Annotators were not pressured to complete annotations, were given a possibility of opting out of categories that they found particularly disturbing, and it was made clear that grading was independent of the completion of the annotation work.

\bibliographystyle{acl_natbib}
\bibliography{arxiv24-fine-granular-trigger-warnings-lit}

\appendix
\clearpage
\onecolumn 
\section{Figures and Tables}\label{appendix-a}

\subsection{Dataset Evaluation}\label{appendix-dataset-eval}

\begin{table*}[h]%
\small
\renewcommand{\tabcolsep}{5pt}%
\renewcommand{\arraystretch}{1}%
\begin{tabular}{@{}l l@{}}
\bf Racism\\
\toprule
\bf Prompt  & Provide a list of racist language that must be avoided in a story for people that are triggered by it. \\
& Return the list in json format.  \\
\midrule
\bf Phrases & nigger,  chink,  spic,  gook,  wetback,  jap,  darkie,  kike,  dago, ngger, chnk, spc, g**k, \\
\bf in Set 1& towelhead, curry muncher, redskin, half-breed, slant eyes \\
\midrule
\bf Phrases & model minority, savage, primitive, barbaric, uncivilized, tribal, blackface, white supremacy, aryan race, \\
\bf in Set 2 & master race, racial purity, colorblind, racism, articulate for, pretty for a, go back to your country, \\
& you're not like other, you're one of the good ones, I don't see color, all lives matter \\
\midrule
\bf Omitted & 
sand nigger, exotic, oriental, reverse racism, nip, speak english \\
\bottomrule
\end{tabular}%
\vspace{2ex}
\begin{tabular}{@{}l l@{}}
\bf Death \\
\toprule
\bf Prompt  & Provide a list of language relating to the death of others, character death, killing, corpses, or coping \\
& with loss or grief that must be avoided in a story for people that are triggered by it. \\
& Return the list in json format. \\
\midrule
\bf Phrases &  mourn, grieve, deal with loss, cope with grief, bury, inter, cremate, come to terms with, kill, murder, \\
\bf in Set 1& slay, weep, slaughter, sacrifice, take a life, end a life, cause death, corpse, dead body, lifeless form, \\
& cadaver, pass away, suffer loss, experience grief, navigate bereavement, struggle with mourning, \\
& manage sorrow, cope with death, die, cease, euthanize \\
\midrule
\bf Phrases & choose to die, end suffering, peacefully pass away, undead, reanimate, come back to life, resurrect, revive, \\
\bf in Set 2& consume flesh, cannibalize, consume brains, devour flesh, tear into, consume blood, drink blood, suck blood, \\
 & cannibalism, necromancy, flesh-eating, blood-sucking, risen from the dead \\

\midrule
\bf Omitted & cry, sob, lament, suffer, comfort, console, remember, honor, respect, accept, process, deal with, face, \\
& confront, put down, put to sleep, remains, lose, deal with deceased, struggle, fight, gasping, gasp, \\ 
& drown, sink, fade, weaken, wane, slip away, pass, depart, succumb, let go, release, end life, attack, \\
& bite, devour, consume, feast, eat, chew, gnaw, rip, tear, destroy, eliminate, terminate, annihilate, \\
& exterminate, rise again, prey upon, feast on, rip apart, consume human flesh, zombify, \\
& reanimate as a zombie, drain blood, vampirize, turn into a vampire, resurrection, zombification, \\
& vampirism, undeadness, death-like state, revival, post-mortem consumption, reanimation, expire \\
\bottomrule
\end{tabular}%
\caption{\warning{Warning Racism and Death.} Keyword selection for two warnings. We selected segments to be annotated based on topical keywords and phrases for each warning. The lists of {\bf Phrases} were generated using {\tt gpt-3.5-turbo-0301} using the provided prompt. We manually removed keywords that better match a different warning ({\it chew, attack, \dots}), frequently lead to false positive matches ({\it let go, depart, lose, \dots}), and which had a large lexical overlap with non-omitted keywords (since our segment retrieval method applies stemming and partial matches). The remaining phrases were split into two sets for the out-of-distribution experiments. The keywords for all warnings can be found in the repository linked in the Abstract.}%
\label{table-appendix-keyword-bootstrapping-example}%
\end{table*} 

\begin{table*}[h]%
\centering%
\small
\renewcommand{\tabcolsep}{4pt}%
\renewcommand{\arraystretch}{.9}%
\begin{tabular}{@{}l cc cc cccc @{}}
\toprule
\multicolumn{1}{@{}l@{}}{\bfseries Warning} & 
\multicolumn{2}{@{}c@{}}{\bfseries Majority Instances} & 
\multicolumn{2}{@{}c@{}}{\bfseries Minority Instances} &
\multicolumn{2}{@{}c@{}}{\bfseries Majority Vote PR} & 
\multicolumn{2}{@{}c@{}}{\bfseries Minority Vote PR}
\\
\cmidrule(l{\tabcolsep}r{\tabcolsep}){2-3}
\cmidrule(l{\tabcolsep}r{\tabcolsep}){4-5}
\cmidrule(l{\tabcolsep}r{\tabcolsep}){6-7}
\cmidrule(l{\tabcolsep}r{\tabcolsep}){8-9}
&
pos & neg & 
pos & neg &
Set 1 & Set 2 &
Set 1 & Set 2 
\\
\midrule
Violence        & 188   & \z\,853   & \z\,363 (35\%)    & \z\,678   & 107 (16\,\%)    & \z81 (22\,\%)   & 205 (30\,\%)  & \z\,158 (43\,\%)  \\
Death           & 114   & \z\,430   & \z\,234 (43\%)    & \z\,310   & \z79 (44\,\%)   & \z35 (10\,\%)   & 122 (69\,\%)  & \z\,112 (31\,\%)  \\
War             & 102   & \z\,725   & \z\,282 (34\%)    & \z\,545   & \z39 (19\,\%)   & \z63 (10\,\%)   & \z97 (48\,\%) & \z\,185 (30\,\%)  \\
Abduction       & 132   & \z\,379   & \z\,273 (53\%)    & \z\,238   & \z59 (23\,\%)   & \z73 (28\,\%)   & 122 (48\,\%)  & \z\,151 (57\,\%)  \\
\midrule
Racism          & 125   & \z\,142   & \z\,185 (69\%)    & \z\z\,82  & \z66 (97\,\%)   & \z59 (30\,\%)   & \z67 (99\,\%) & \z\,118 (59\,\%)  \\
Homophobia      & 138   & \z\,175   & \z\,216 (69\%)    & \z\z\,97  & \z52 (40\,\%)   & \z86 (47\,\%)   & \z77 (60\,\%) & \z\,139 (76\,\%)  \\
Misogyny        & \z81  & \z\,296   & \z\,179 (47\%)    & \z\,198   & \z57 (38\,\%)   & \z24 (11\,\%)   & \z90 (59\,\%) & \z\,\z89 (40\,\%) \\
Ableism         & \z87  & \z\,168   & \z\,169 (66\%)    & \z\z\,86  & \z43 (27\,\%)   & \z44 (47\,\%)   & \z92 (56\,\%) & \z\,\z77 (83\,\%) \\
\midrule
Total           & 966   & 3,169     & 1,901 (46\%)      & 2,234     & 501 (28\,\%)    & 465 (20\,\%)    & 872 (48\,\%)  & 1,029 (44\,\%)    \\
\bottomrule
\end{tabular}%
\caption{The number and ratio of positive and negative passages under majority and minority vote aggregation and the number and ratio (PR) of positively annotated examples across warnings. Shown are the PRs assuming Majority Vote (2 or more positives) and Minority Vote (1 or more positives), depending on if the keyword (used to retrieve the passage) was in set 1 or 2.}%
\label{table-appendix-extended-vote-distribution}%
\end{table*}

\clearpage
\subsection{Classification Evaluation}\label{appendix-classification-eval}
\newcommand{\tabhead}{
\multicolumn{1}{@{}l@{}}{\bfseries Mode} & 
\multicolumn{2}{@{}c}{\bfseries Violence} & 
\multicolumn{2}{@{}c}{\bfseries Death} & 
\multicolumn{2}{@{}c}{\bfseries War} & 
\multicolumn{2}{@{}c}{\bfseries Abduction} & 

\multicolumn{2}{@{}c}{\bfseries Racism} & 
\multicolumn{2}{@{}c}{\bfseries Homoph.} & 
\multicolumn{2}{@{}c}{\bfseries Misogyny} & 
\multicolumn{2}{@{}c}{\bfseries Ableism} & 
\multicolumn{2}{@{}c}{\bfseries Mean}
\\
\cmidrule(l{\tabcolsep}r{\tabcolsep}){2-3}
\cmidrule(l{\tabcolsep}r{\tabcolsep}){4-5}
\cmidrule(l{\tabcolsep}r{\tabcolsep}){6-7}
\cmidrule(l{\tabcolsep}r{\tabcolsep}){8-9}
\cmidrule(l{\tabcolsep}r{\tabcolsep}){10-11}
\cmidrule(l{\tabcolsep}r{\tabcolsep}){12-13}
\cmidrule(l{\tabcolsep}r{\tabcolsep}){14-15}
\cmidrule(l{\tabcolsep}r{\tabcolsep}){16-17}
\cmidrule(l{\tabcolsep}r{\tabcolsep}){18-19}
&
Acc & PR & 
Acc & PR & 
Acc & PR & 
Acc & PR & 

Acc & PR & 
Acc & PR & 
Acc & PR & 
Acc & PR & 
Acc & PR 
}
\begin{table*}[ht]%
\small
\renewcommand{\tabcolsep}{3.5pt}%
\renewcommand{\arraystretch}{.9}%
{\bf(a)} In-Distribution with Minority Voting
\\
\begin{tabular}{@{}l cc cc cc cc | cc cc cc cc | cc @{}}
\toprule
\tabhead 
\\
\midrule
Binary	      & \bf0.80 & 0.44	& \bf0.81 & 0.45	& 0.70 & 0.42	& 0.76 & 0.53	& 0.74 & 0.57	& 0.64 & 0.74	& 0.63 & 0.46	& 0.59 & 0.83 & 0.71 & 0.56 \\
Binary+	      & 0.77 & 0.43	& 0.74 & 0.48	& 0.71 & 0.43	& 0.67 & 0.60	& 0.68 & 0.38	& 0.72 & 0.58	& 0.65 & 0.38	& 0.71 & 0.55 & 0.71 & 0.48 \\
Multiclass	  & 0.78 & 0.28	& 0.78 & 0.29	&\bf 0.77 & 0.27	& \bf0.86 & 0.36	& \bf0.81 & 0.32	& \bf0.91 & 0.41	& \bf0.79 & 0.28	& \bf0.90 & 0.40 & \bf0.82 & 0.32 \\
Multilabel	  & 0.58 & 0.73	& 0.48 & 0.90	& 0.56 & 0.82	& 0.68 & 0.76	& 0.51 & 0.77	& 0.55 & 0.93	& 0.51 & 0.86	& 0.68 & 0.75 & 0.57 & 0.82 \\
Multilabel+	  & 0.61 & 0.62	& 0.48 & 0.86	& 0.54 & 0.78	& 0.67 & 0.73	& 0.63 & 0.69	& 0.50 & 0.00	& 0.56 & 0.80	& 0.70 & 0.71 & 0.59 & 0.65 \\
\midrule
GPT 3.5	      & 0.76 & 0.53	& \bf0.72 & 0.67	& 0.66 & 0.77	& 0.78 & 0.58	& 0.65 & 0.68	& 0.70 & 0.74	& 0.59 & 0.76	& 0.70 & 0.60 & \bf0.70 & 0.67 \\
GPT 4 	      & \bf0.78 & 0.68	& 0.57 & 0.91	& 0.67 & 0.74	& \bf0.83 & 0.45	& 0.69 & 0.65	& \bf0.71 & 0.74	& 0.60 & 0.70	& \bf0.75 & 0.60 & \bf0.70 & 0.68 \\
Mistral 7B	  & 0.74 & 0.68	& 0.67 & 0.78	& \bf0.68 & 0.80	& 0.73 & 0.69	& 0.64 & 0.65	& 0.63 & 0.83	& 0.62 & 0.64	& 0.67 & 0.59 & 0.67 & 0.71 \\
Mixtral 8x7B  & \bf0.78 & 0.42	& 0.70 & 0.56	& 0.62 & 0.68	& 0.76 & 0.42	& \bf0.67 & 0.48	& \bf0.71 & 0.55	& \bf0.64 & 0.57	& 0.70 & 0.36 & \bf0.70 & 0.51 \\
Llama 7B      & 0.70 & 0.32	& 0.68 & 0.71	& 0.63 & 0.55	& 0.64 & 0.45	& 0.65 & 0.57	& 0.63 & 0.68	& 0.56 & 0.50	& 0.56 & 0.21 & 0.63 & 0.50 \\
Llama 13B     & 0.68 & 0.67	& 0.62 & 0.82	& 0.66 & 0.66	& 0.64 & 0.76	& 0.60 & 0.77	& 0.65 & 0.75	& 0.54 & 0.77	& 0.62 & 0.75 & 0.63 & 0.74 \\
\midrule
Mean	      & 0.72 & 0.53	& 0.66 & 0.68	& 0.65 & 0.63	& 0.73 & 0.58	& 0.66 & 0.59	& 0.67 & 0.63	& 0.61 & 0.61	& 0.69 & 0.58	\\
\bottomrule
\end{tabular}
\par
{\bf(c)} In-Distribution with Majority Voting
\rule{0em}{3ex} \\
\begin{tabular}{@{}l cc cc cc cc | cc cc cc cc | cc @{}}
\toprule
\tabhead 
\\
\midrule
Binary	      & \bf0.77 & 0.32	& \bf0.81 & 0.32	& 0.68 & 0.18	& 0.57 & 0.18	& 0.74 & 0.47	& 0.71 & 0.43	& 0.57 & 0.14	& 0.55 & 0.24 & 0.68 & 0.28 \\
Binary+	      & 0.75 & 0.32	& 0.79 & 0.34	& 0.61 & 0.12	& 0.63 & 0.24	& 0.73 & 0.34	& 0.67 & 0.42	& 0.60 & 0.15	& 0.59 & 0.26 & 0.67 & 0.27 \\
Multiclass	  & 0.70 & 0.20	& 0.73 & 0.23	& 0.64 & 0.14	& 0.65 & 0.15	& \bf0.81 & 0.30	& \bf0.84 & 0.34	& \bf0.63 & 0.13	& \bf0.71 & 0.22 & \bf0.71 & 0.21 \\
Multilabel	  & 0.71 & 0.57	& 0.49 & 0.86	& 0.59 & 0.78	& \bf0.72 & 0.71	& 0.52 & 0.82	& 0.49 & 0.94	& 0.57 & 0.79	& 0.55 & 0.77 & 0.58 & 0.78 \\
Multilabel+	  & 0.74 & 0.41	& 0.62 & 0.70	& \bf0.73 & 0.55	& \bf0.72 & 0.59	& 0.61 & 0.71	& 0.50 & 0.00	& 0.62 & 0.62	& 0.58 & 0.78 & 0.64 & 0.54 \\
\midrule
GPT 3.5	      & \bf0.84 & 0.57	& 0.71 & 0.79	& 0.70 & 0.78	& 0.74 & 0.74	& 0.68 & 0.73	& 0.63 & 0.86	& 0.66 & 0.84	& 0.66 & 0.70 & 0.70 & 0.75 \\
GPT 4	      & 0.80 & 0.70	& 0.60 & 0.90	& 0.69 & 0.81	& 0.76 & 0.66	& \bf0.72 & 0.66	& 0.62 & 0.87	& 0.68 & 0.80	& 0.68 & 0.72 & 0.70 & 0.76 \\
Mistral 7B	  & 0.74 & 0.70	& 0.62 & 0.87	& 0.64 & 0.82	& 0.68 & 0.82	& 0.70 & 0.66	& 0.54 & 0.94	& 0.68 & 0.75	& 0.66 & 0.73 & 0.66 & 0.79 \\
Mixtral 8x7B  & 0.80 & 0.49	& \bf0.76 & 0.68	& \bf0.72 & 0.71	& \bf0.77 & 0.54	& \bf0.72 & 0.53	& \bf0.73 & 0.70	& \bf0.71 & 0.70	& \bf0.74 & 0.56 & \bf0.74 & 0.61 \\
Llama 7B	  & 0.73 & 0.40	& 0.63 & 0.79	& 0.71 & 0.57	& 0.65 & 0.57	& \bf0.72 & 0.56	& 0.67 & 0.77	& 0.65 & 0.66	& 0.60 & 0.34 & 0.67 & 0.58 \\
Llama 13B	  & 0.70 & 0.64	& 0.58 & 0.88	& 0.66 & 0.69	& 0.64 & 0.80	& 0.62 & 0.78	& 0.58 & 0.92	& 0.65 & 0.83	& 0.58 & 0.82 & 0.63 & 0.80 \\
\midrule
Mean	      & 0.75 & 0.48	& 0.67 & 0.67	& 0.67 & 0.56	& 0.69 & 0.55	& 0.69 & 0.60	& 0.64 & 0.65	& 0.64 & 0.58	& 0.63 & 0.56 \\
\bottomrule	
\end{tabular}
\par
{\bf(c)} Out-of-Distribution with Minority Voting
\rule{0em}{3ex} \\
\begin{tabular}{@{}l cc cc cc cc | cc cc cc cc | cc @{}}
\toprule
\tabhead 
\\
\midrule
Binary	      & \bf0.76 & 0.36	& 0.72 & 0.41	& 0.66 & 0.62	& \bf0.76 & 0.53	& \bf0.50 & 1.00	& \bf0.62 & 0.22	& 0.56 & 0.63	& 0.48 & 0.46 & \bf0.63 & 0.53 \\
Binary+	      & \bf0.76 & 0.40	& \bf0.73 & 0.44	& 0.68 & 0.59	& 0.66 & 0.49	& 0.47 & 0.20	& 0.57 & 0.30	& 0.55 & 0.14	& 0.43 & 0.40 & 0.61 & 0.37 \\
Multiclass	  & 0.71 & 0.21	& 0.55 & 0.05	& \bf0.77 & 0.27	& 0.71 & 0.21	& \bf0.50 & 0.00	& 0.50 & 0.00	& \bf0.57 & 0.08	& \bf0.55 & 0.11 & 0.61 & 0.12 \\
Multilabel	  & 0.59 & 0.57	& 0.62 & 0.48	& 0.61 & 0.68	& 0.63 & 0.65	& 0.47 & 0.28	& 0.43 & 0.48	& 0.49 & 0.36	& 0.47 & 0.56 & 0.54 & 0.50 \\
Multilabel+	  & 0.67 & 0.45	& 0.66 & 0.22	& 0.69 & 0.64	& 0.70 & 0.67	& \bf0.50 & 0.02	& 0.50 & 0.00	& \bf0.57 & 0.18	& 0.47 & 0.26 & 0.60 & 0.31 \\
\midrule
GPT 3.5	      & 0.73 & 0.55	& \bf0.71 & 0.58	& 0.71 & 0.76	& \bf0.80 & 0.55	& 0.58 & 0.62	& \bf0.68 & 0.77	& 0.55 & 0.68	& \bf0.79 & 0.61 & \bf0.70 & 0.64 \\
GPT 4	        & \bf0.79 & 0.66	& 0.56 & 0.90	& \bf0.74 & 0.73	& 0.74 & 0.51	& \bf0.61 & 0.50	& 0.63 & 0.85	& 0.59 & 0.53	& 0.72 & 0.76 & 0.67 & 0.68 \\
Mistral 7B	  & 0.72 & 0.66	& 0.63 & 0.76	& 0.71 & 0.78	& 0.69 & 0.67	& 0.58 & 0.49	& 0.55 & 0.92	& 0.61 & 0.43	& 0.72 & 0.56 & 0.65 & 0.66 \\
Mixtral 8x7B	& 0.74 & 0.43	& 0.64 & 0.48	& \bf0.74 & 0.70	& 0.75 & 0.39	& 0.51 & 0.34	& 0.65 & 0.62	& \bf0.63 & 0.36	& 0.76 & 0.38 & 0.68 & 0.46 \\
Llama 7B	    & 0.64 & 0.41	& 0.60 & 0.64	& 0.67 & 0.51	& 0.69 & 0.45	& 0.48 & 0.44	& \bf0.68 & 0.71	& 0.52 & 0.30	& 0.51 & 0.29 & 0.60 & 0.47 \\
Llama 13B	    & 0.65 & 0.65	& 0.57 & 0.84	& \bf0.74 & 0.67	& 0.66 & 0.69	& 0.56 & 0.67	& 0.65 & 0.82	& 0.52 & 0.61	& 0.61 & 0.81 & 0.62 & 0.72 \\
\midrule
Mean	            & 0.70 & 0.48	& 0.64 & 0.53	& 0.70 & 0.63	& 0.71 & 0.53	& 0.52 & 0.41	& 0.59 & 0.52	& 0.56 & 0.39	& 0.59 & 0.47 \\
\bottomrule
\end{tabular}
\par
{\bf(d)} Out-of-Distribution with Majority Voting
\rule{0em}{3ex} \\
\begin{tabular}{@{}l cc cc cc cc | cc cc cc cc | cc @{}}
\toprule
\tabhead 
\\
\midrule
Binary	      & 0.68 & 0.25	& 0.59 & 0.10	& 0.64 & 0.28	& 0.56 & 0.26	& \bf0.50 & 1.00	& 0.53 & 0.09	& \bf0.56 & 0.43	& 0.49 & 0.08 & 0.57 & 0.31 \\
Binary+	      & 0.70 & 0.28	& 0.68 & 0.30	& \bf0.65 & 0.30	& 0.61 & 0.44	& 0.48 & 0.10	& 0.52 & 0.09	& 0.50 & 0.02	& 0.46 & 0.08 & 0.58 & 0.20 \\
Multiclass	  & 0.62 & 0.12	& 0.57 & 0.08	& 0.62 & 0.12	& \bf0.69 & 0.20	& \bf0.50 & 0.00	& 0.50 & 0.00	& 0.50 & 0.00	& \bf0.50 & 0.00 & 0.57 & 0.07 \\
Multilabel	  & 0.66 & 0.52	& \bf0.71 & 0.32	& 0.64 & 0.57	& 0.65 & 0.70	& 0.46 & 0.23	& \bf0.56 & 0.45	& 0.51 & 0.26	& \bf0.50 & 0.36 & 0.59 & 0.43 \\
Multilabel+	  & \bf0.72 & 0.35	& 0.70 & 0.28	& \bf0.65 & 0.61	& 0.67 & 0.63	& \bf0.50 & 0.02	& 0.50 & 0.00	& 0.54 & 0.13	& 0.49 & 0.22 & \bf0.60 & 0.28 \\
\midrule
GPT 3.5	      & \bf0.79 & 0.66	& 0.76 & 0.72	& 0.66 & 0.82	& 0.72 & 0.78	& \bf0.60 & 0.60	& 0.62 & 0.86	& 0.68 & 0.69	& 0.72 & 0.76 & 0.69 & 0.74 \\
GPT 4	        & 0.71 & 0.80	& 0.59 & 0.90	& 0.68 & 0.80	& 0.71 & 0.74	& 0.59 & 0.51	& 0.58 & 0.92	& 0.67 & 0.61	& 0.63 & 0.84 & 0.65 & 0.77 \\
Mistral 7B	  & 0.68 & 0.76	& 0.66 & 0.84	& 0.67 & 0.80	& 0.65 & 0.85	& 0.54 & 0.54	& 0.55 & 0.94	& 0.68 & 0.54	& 0.71 & 0.68 & 0.64 & 0.74 \\
Mixtral 8x7B	& 0.78 & 0.55	& \bf0.77 & 0.63	& 0.68 & 0.72	& \bf0.73 & 0.61	& 0.54 & 0.32	& \bf0.71 & 0.70	& \bf0.70 & 0.44	& \bf0.74 & 0.53 & \bf0.71 & 0.56 \\
Llama 7B	    & 0.72 & 0.42	& 0.62 & 0.74	& 0.71 & 0.52	& 0.66 & 0.62	& 0.59 & 0.52	& 0.65 & 0.75	& 0.61 & 0.34	& 0.57 & 0.30 & 0.64 & 0.53 \\
Llama 13B	    & 0.65 & 0.77	& 0.63 & 0.86	& \bf0.72 & 0.64	& 0.65 & 0.84	& 0.55 & 0.70	& 0.60 & 0.89	& 0.65 & 0.68	& 0.57 & 0.85 & 0.63 & 0.78 \\
\midrule
Mean	            & 0.70 & 0.50	& 0.66 & 0.52	& 0.67 & 0.56	& 0.66 & 0.61	& 0.53 & 0.41	& 0.58 & 0.52	& 0.60 & 0.38	& 0.58 & 0.43	\\
\bottomrule  
\end{tabular}%
\caption{Mean accuracy (Acc) and positive rate (PR) across five folds of all 11 tested models across all 8 tested trigger warnings. The best model by accuracy is marked bold for each warning. The bar charts in Figure~\ref{figure-appendix-model-performance-mean-barchats}(b) show the mean across each model with CI.}%
\label{table-appendix-classification-results}%
\end{table*}
\begin{figure}[h]
\centering
\includegraphics[scale=0.9]{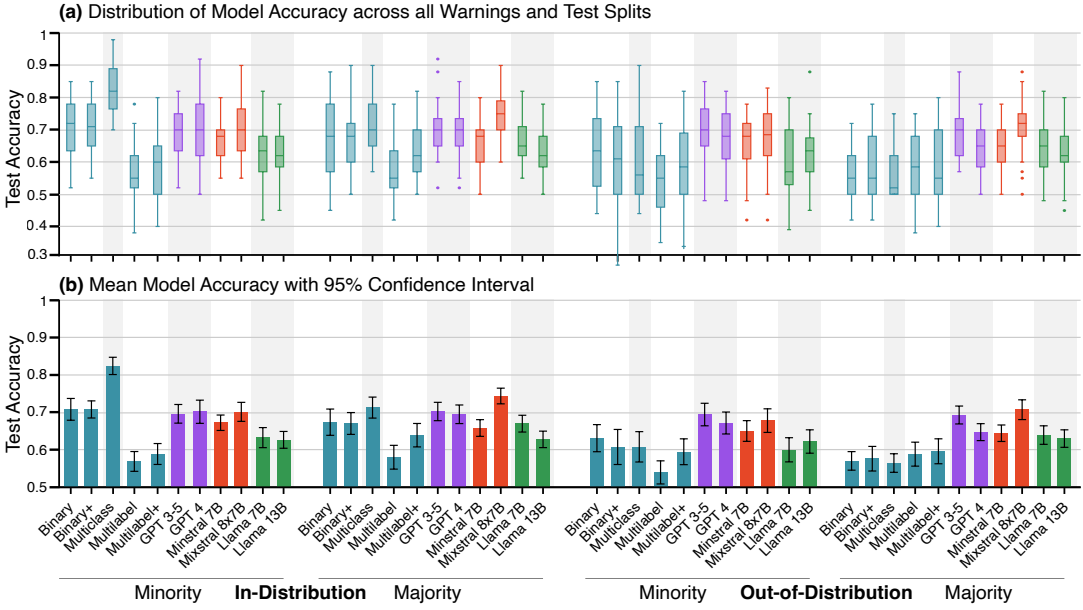}
\caption{Distribution and mean of model accuracy across all tested models. The coloring is for visual ease only.}\label{figure-appendix-model-performance-mean-barchats}  
\end{figure}



\clearpage
\twocolumn
\section{Model Training}\label{appendix-b}

All fine-tuned models are based on a `roberta-base' checkpoint that was fine-tuned for fan fiction using masked language modeling (cf.~\ref{fanbert}). We conducted a parameter sweep for each fine-tuned model on a 6th fold (cf.~\ref{parameter-sweep}). All fine-tuning models were trained on a single A100 40GB. All generative LLMs were prompted using the instructions shown in Figure~\ref{figure-instruction-and-prompt-parts}. Section~\ref{prompt-ablation} describes the prompt and our ablation study in detail. All used models are explained in detail in Section~\ref{model-implementation}.

\subsection{Language Modeling Fine-tuning of RoBERTa for Fan Fiction}\label{fanbert}

We fine-tuned the `roberta-base' checkpoint on fan fiction documents via masked language modeling using Huggingface's `Trainer' routine. As data, we extracted all English fan fiction documents from WTWC-22 that were marked as recommended for model training, i.e. where a trigger warning was assigned and where overly long, short, or unpopular works were removed (\citet{wiegmann:2023a} list the precise parameters). All documents were split into training examples of ca. 450-500 words, respecting sentence and paragraph boundaries. We trained the checkpoint on the resulting 19 million examples for ca. 470,000 steps with a batch size of 32, the point of loss convergence on a 10,000 example hold-out validation set. As parameters, we used largely the standard parameters of the `Trainer', with a random masking function, masking probability of 0.2, and an initial learning rate of 2e-5. 

\subsection{Fine-tuning Parameter Sweep}\label{parameter-sweep}
We conducted a grid-based parameter sweep on the validation sample (cf. Section~\ref{experiment-datasets}) for all fine-tuning strategy (binary, multi-label, multi-class, with and without extended training data) across three dimensions: in vs out-of-distribution, minority vs. majority vote aggregation, and learning rate within $(1e-5, 2e-5, 5e-5)$. All models were trained for 20 epochs. The parameters did not vary between warnings. Table~\ref{table-appendix-parameter-sweep} shows the final parameter settings each model was trained on. 

\begin{table}[h]%
\centering%
\small
\renewcommand{\tabcolsep}{5pt}%
\renewcommand{\arraystretch}{1}%
\begin{tabular}{@{}ll ll r @{}}
\toprule
\bf Strategy & \bf Dist. & \bf  Vote Agg. & \bf LR & \bf Acc. \\
\midrule
binary & ood    & majority & 1e-5lr & \bf 0.62 \\
binary & ood    & minority & 2e-5lr & \bf 0.64 \\
binary & id     & majority & 1e-5lr & \bf 0.68 \\
binary & id     & minority & 5e-5lr & 0.74 \\
\midrule
binary extended & ood   & minority & 2e-5lr & \bf 0.62 \\
binary extended & ood   & majority & 2e-5lr & 0.62 \\
binary extended & id    & majority & 2e-5lr & 0.64 \\
binary extended & id    & minority & 1e-5lr & 0.71 \\
\midrule
multi-label & ood    & majority & 2e-5lr & 0.58 \\
multi-label & ood    & minority & 5e-5lr & 0.57 \\
multi-label & id     & majority & 2e-5lr & 0.59 \\
multi-label & id     & minority & 1e-5lr & 0.57 \\
\midrule
multi-label extended & ood  & majority & 2e-5lr & \bf 0.62 \\
multi-label extended & ood  & minority & 2e-5lr & 0.58 \\
multi-label extended & id   & majority & 1e-5lr & 0.66 \\
multi-label extended & id   & minority & 1e-5lr & 0.57 \\
\midrule
multi-class & ood   & majority & 2e-5lr & 0.59 \\
multi-class & ood   & minority & 5e-5lr & \bf 0.64 \\
multi-class & id    & majority & 2e-5lr & \bf 0.68 \\
multi-class & id    & minority & 5e-5lr & \bf 0.84 \\
\bottomrule
\end{tabular}%
\caption{Overview of the different settings, the final parameter choices for each model, and the accuracy score (mean across all warnings) of that setting. The different tested learning rates varied between 0.01--0.09.}%
\label{table-appendix-parameter-sweep}%
\end{table} 
    
\subsection{Prompt Ablation}\label{prompt-ablation}

Figure~\ref{figure-instruction-and-prompt-parts} shows the 5 parts of the instructions we used for both the annotators and as the prompt for the few-shot models: {\it Instruction}, {\it Passage}, {\it Persona}, {\it Definition}, and {\it Demonstrations} (one positive and one negative in that order). We counted model responses as positive when they contained `yes' within the first 5 response tokens and negative if they contained `no'

For the prompt ablation, we queried all generative LLMs with open weights (excluding GPT) across all 4 test settings (aggregation and distribution), averaged the scores across all settings per model, and averaged the scores across all models to determine the best prompt template. We tested eight prompt template variations (cf. Table~\ref{table-prompt-ablation}). Each variation started with {\it Instruction} and {\it Passage}, followed by all combinations of {\it Persona}, {\it Definition}, and {\it Demonstrations}, including neither. We manually selected the demonstrations that we rated as clear and representative for the respective trigger out of all unanimously annotated passages. We selected the prompt with all parts, which had with the highest average accuracy on the validation sample across all models and was also the one shown to annotators.

\begin{table*}[h]%
\centering%
\small
\renewcommand{\tabcolsep}{2pt}%
\renewcommand{\arraystretch}{1}%
\begin{tabular}{@{}lll cc cc c @{}}
\toprule

\multicolumn{3}{@{}c@{}}{\bfseries Prompt} &  \bf Mistral 7B &  \bf Mixtral 8x7B & \bf Llama 7B & \bf Llama 13B & \bf Mean Acc. \\
\midrule
\multicolumn{3}{@{}l@{}}{Instruction and Passage} & 0.72     & 0.74       &  0.52     & 0.50     & 0.62 \\
+ Persona &              &                        & 0.71     & 0.74       &  0.57     & 0.57     & 0.65 \\
          & + Definition &                        & 0.72     & \bf 0.75   &  0.54     & 0.58     & 0.65 \\
          &              & + Demonstrations       & 0.69     & 0.70       &  0.58     & \bf 0.64 & 0.65 \\
\midrule                                  
+ Persona & + Definition &                        & \bf 0.73 & \bf 0.75   &  0.57     & 0.55     & 0.65 \\
+ Persona &              & + Demonstrations       & 0.68     & 0.74       &  0.64     & \bf 0.64 & \bf 0.68 \\
          & + Definition & + Demonstrations       & 0.68     & \bf 0.75   &  0.67     & 0.63     & \bf 0.68 \\
+ Persona & + Definition & + Demonstrations       & 0.68     & \bf 0.75   &  \bf 0.68 & 0.59     & \bf 0.68 \\
\bottomrule

\end{tabular}%
\caption{Overview of the accuracy on the validation sample of different candidate prompts across the tested models. All prompts ended with {\it Instruction} and {\it Passage} but varied in other parts.}%
\label{table-prompt-ablation}%
\end{table*} 

\end{document}